\documentclass[runningheads]{llncs}
\usepackage{graphicx}
\usepackage{times}
\usepackage{latexsym}
\usepackage[utf8]{inputenc}
\usepackage{graphicx}
\usepackage[]{algorithm2e}
\usepackage{multirow}
\usepackage{graphicx}
\usepackage{caption}
\usepackage{hhline}
\usepackage{amsmath}
\usepackage{booktabs}
\usepackage{enumitem}
\usepackage{hyperref}
\usepackage{enumitem}
\usepackage{cite}
\usepackage{url}
\usepackage{microtype}

\begin{document}

\title{SciBERTSUM: Extractive Summarization for Scientific Documents}

\author{Athar Sefid  \inst{1} \and
C Lee Giles \inst{1,2}
}

%

\institute{Computer Science and Engineering, Pennsylvania State University, University, PA, 16802, USA \and
Information Sciences \& Technology Department, Pennsylvania State University, University Park, PA, 16802, USA \\
\email{atharsefid@gmail.com, clg20@psu.edu}
}

\maketitle      

\begin{abstract}

The summarization literature focuses on the summarization of news articles. The news articles in the CNN-DailyMail are relatively short documents with about 30 sentences per document on average. We introduce SciBERTSUM, our summarization framework designed for the summarization of long documents like scientific papers with more than 500 sentences. SciBERTSUM extends BERTSUM to long documents by 1) adding a section embedding layer to include section information in the sentence vector and 2) applying a sparse attention mechanism where each sentences will attend locally to nearby sentences and only a small number of sentences attend globally to all other sentences. We used slides generated by the authors of scientific papers as reference summaries since they contain the technical details from the paper. The results show the superiority of our model in terms of ROUGE scores. \footnote{The code is available at \hyperlink{https://github.com/atharsefid/SciBERTSUM}{https://github.com/atharsefid/SciBERTSUM}.}
\end{abstract}

\section{Introduction}

Automatic  summarization frameworks condense an input document into shorter text consisting of the main points in that document. Neural networks have achieved state of the art results for both paradigms of abstractive summarization\cite{see-etal-2017-get, celikyilmaz-etal-2018-deep} and extractive summarization \cite{nallapati2017summarunner, narayan2018ranking}. While extractive models are factually more consistent with the content in the input document, abstractive models can be novel and less redundant.
Most of the existing methods are used on news datasets \cite{nallapati-etal-2016-abstractive,grusky-etal-2018-newsroom} where the input document is relatively short and normally less than 30 sentences long. 
Summarization of long documents such as scientific papers is different from a short article summarization since it requires more memory and computational power to encode the full document and model the relationship between the sentences. 

Natural language processing applications have completely been revolutionized with the advent of pre-trained models. 
 Pre-trained language models are easy to incorporate and don’t require much-labeled data to deal with, which makes it appropriate for many problems such as prediction, transfer learning, and feature extraction.
 Bidirectional Encoder Representations from Transformers (BERT) \cite{devlin-etal-2019-bert} have combined both word and sentence representations into a single very large Transformer \cite{NIPS2017_7181}. This has shown superior results on many NLP tasks such as question answering and text generation. 
BERT was trained on large amounts of data with the objective of predicting the masked tokens and the next sentence and it can be fine-tuned for various task-specific objectives \cite{mosbach2020stability}.

Language models such as BERT \cite{devlin-etal-2019-bert} or SciBERT \cite{beltagy-etal-2019-scibert} have improved many language based tasks, especially with SciBERT for science related documents. The impact of BERT on extractive summarization was due to BERTSUM. BERTSUM extended BERT from a two-sentence language model to one that covers all sentences in a document. BERTSUM model with a full attention layer can capture the document-level features. However, full attention is not efficient for the summarization of long documents such as scientific papers which have more than 500 sentences. Here we propose an extractive transformer based summarizer for longer documents such as scientific articles with multiple sections.

The contributions  of our model are:
 \begin{itemize}
 \item Design a section embedding layer to the embedding module of BERTSUM where all tokens in the same section are embedded with the same embedding token. This is crucial for the embedding of long documents with multiple sections in a hierarchical structure. 
 
 \item Employ a sparse inter-sentence attentional model with local and global attention schemes where each sentence will attend locally to nearby sentences and some random sentences attend globally to all other sentences in the document. 
 
 \item Devise summarization modules for scientific articles using the presentation slides as the ground-truth summaries. The slides contain the technical details from the paper and usually follow the structure of the papers.
 
 \end{itemize}
 
    
\section{Related Work}

\subsection{Summarization}

We believe summarizing scientific articles is more challenging than summarizing generic text since such articles have a hierarchical structure \cite{IBRAHIMALTMAMI2020}. They contain technical terms and formulas \cite{yasunaga2019scisummnet}, and much valuable content can be embedded in figures, tables, and algorithms \cite{10.1145/2094072.2094075}.  

Scientific article summarization has been less investigated compared to news articles summarization \cite{cheng-lapata-2016-neural,see-etal-2017-get}. This seems to be mainly due to the lack of training data for full scientific articles. Types of reference summaries for scientific articles are:
\begin{itemize}[leftmargin=*]
    \item Abstract: Most of the traditional summarization methods use the abstract as the reference summary of the paper. However, abstracts are extremely compressed versions of a papers and usually do not have enough space to include all of the contributions \cite{elkiss2008blind}.
    \item Citation-based: These types of summaries integrate the authors' highlights in the abstract of the paper with citation context of the citing papers which in some ways reflects the impact in the paper of the research community \cite{yasunaga2019scisummnet}. 
    \item Speaker Transcript: Many conference proceedings/ workshops require the authors to verbally present their work. TalkSum \cite{lev2019talksumm} uses the transcript of these presentations as a summary of the scientific article. However, the transcripts in the TalkSum data set are often noisy and can not be readily used as reference summaries.
\end{itemize}

\textbf{Presentation slides for a paper are a different class of summaries that intend to cover in some way the important content of the entire paper, sometimes section by section}. They contain the main highlights and also valuable images/tables. They are not as noisy as speaker transcripts and are becoming more available as more conferences are providing slides that go with their papers. We used the PS5k dataset \cite{sefid-etal-2021-extractive, sefid2021slidegen} to build our summarizer.



\subsection{Transformer Based Summarization}

Pre-trained language models such as BART \cite{lewis-etal-2020-bart}
produce state-of-the-art results on the summarization tasks. However, they are often used on short news articles such as XSum \cite{xsum-emnlp} or CNN-DailyMail \cite{nallapati-etal-2016-abstractive} datasets. These models are not designed for scientific articles and their space/computational complexity grows quadratically  with the size of the input.  

HIBERT \cite{zhang2019hibert} is an extractive summarizer that learns context aware sentence representations using multiple layers of transformers. Here, 15\% of the full sentences are masked (replaced with a single [mask] token) with the goal to predict the sentence embedding of the masked sentences. BERTSUM \cite{liu-lapata-2019-text} is another BERT style extractive summarizer that extends BERT to multiple sentences by expanding the positional embedding and using interval segmentation embeddings to distinguish multiple sentences within a document. Sotudeh et al. \cite{sotudeh-gharebagh-etal-2020-guir} added section information to the objective function of BERTSUM so it could
optimize both 
the sentence prediction and section prediction tasks in a multi-task setting. However, most of these transformer-based extractive summarizers do not scale for long documents with thousands of tokens nor can they be applied to many full scientific documents.

\section{Method - SciBERTSUM}

Most of the previous language models such as BERT are employed as encoders for short pieces of text usually covering at most 2 sentences. The summarization task besides other NLP tasks (e.g. predicting entailment relationship, question answering) requires a wide coverage of the full document containing multiple sections and many sentences. 
We propose a document encoder based on BERT. Our encoder model will help build sentence representations by stacking multiple transformer layers on top of sentence vectors to model the inter sentence relations in the full document.

Our SciBERTSUM model is an extension of BERTSUM and can generate sentence embeddings for all sentences in a full document with multiple sections. Our model applies a linear sparse attention mechanism between sentences to represent inter sentence relations and it outperforms BERTSUM on our dataset.

\section{Language Model Architecture}

To explain the architecture of our language model, we first explain how we generate the sentence embeddings by adding section information to sentences and then we explain how our sparse attention mechanism helps us process the full document efficiently.

\subsection{Embedding Layer}

The embedding layer of BERT \cite{devlin-etal-2019-bert} applies the byte-pair encoding to tokenize the text. It adds [CLS] tokens to the beginning of the sequences and [SEP] tokens to separate the first and second sentences in the sequence. The embedding representation of the [CLS] token is the representation of the full sequence and is used for sentence classification tasks.

BERT combines 1) Semantic embedding (the meaning of the token) 2) Positional embedding (the position of the token in the sequence), 3) Segmentation embedding ( the embedding layer to distinguish the first and the second sentence in the sequence) to form the embedding of a token in a sequence.

\begin{figure*}[htbp]
\centerline{\includegraphics[height=2.5cm, width=0.8\textwidth]{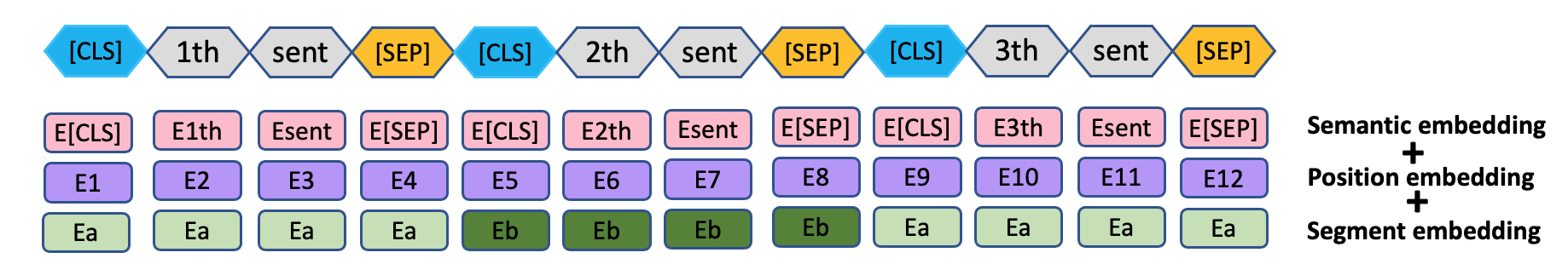}}
\caption[BERTSUM architecture ]{BERTSUM architecture covering multiple sentence. Each sentence has a [CLS] token at the beginning.}
\label{bertsum_architecture}
\end{figure*}

BERTSUM \cite{liu-lapata-2019-text} extends BERT to multiple sentences by adding [CLS] tokens to the beginning of all sentences. It changes the segmentation embedding to distinguish odd and even sentences.
The embedding model is depicted in Figure \ref{bertsum_architecture}. The green boxes are the segmentation embeddings. Light greens are the embedding for odd sentences and dark green boxes are for even sentences. BERTSUM extended the positional embedding of BERT beyond 512 tokens to cover all tokens of the input document.

The sentence embeddings  are the embedding of the [CLS] tokens which are the  combination of semantic, segment, and position embeddings. The Positional encoding is the sinusoid embedding from Vaswani et. al \cite{NIPS2017_7181}. 

\begin{figure*}[htbp]
\centerline{
\includegraphics[height=6cm,width=1.03\textwidth]{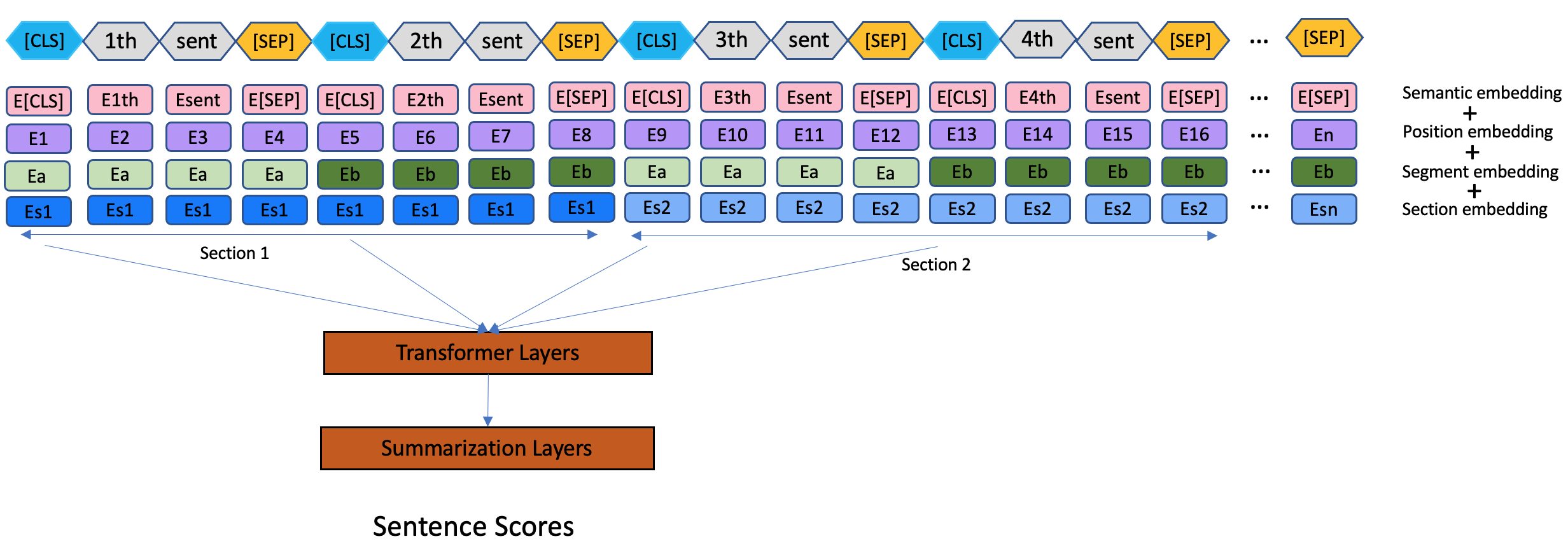}
}
\caption{SciBERTSUM architecture}
\label{SciBERTsum_architecture}
\end{figure*}

Long documents especially long scholarly articles contain multiple sections. The section of the document is important in the selection of salient sentences. For instance, the sentences in the `\textit{acknowledgment}' section are less important compared to other sections like `\textit{abstract}' or `\textit{results}'.  
We enhance BERTSUM by adding section embedding as shown in Figure \ref{SciBERTsum_architecture}. The sentence embeddings ($E_{sents})$ are the combination of the section, semantic, position, and segmentation embeddings. The section embeddings are the blue boxes in Figure \ref{SciBERTsum_architecture}. All of the tokens of the sentences in the first section are embedded by dark blue and the tokens of sentences in the second section are embedded in light blue. 
Each section has the same segmentation embedding as in BERTSUM.

\begin{equation}
   E_{sents} = Semantic + Position  + Segment + Section
\end{equation}

 To overcome the memory issue where we cannot load the full document with the full position embedding in the memory. We get the sentence vectors section by section. We can load the maximum 3072 tokens to the memory based on experiments on an Nvidia GPU with 11,019 MiB memory capacity. 

\subsection{Attention Mechanism }

BERTSUM applies multiple layers of full dot-product attention. However, applying full attention on the sentence vectors is expensive where the number of sentences in documents is large. Scientific documents can have more than 500 sentences. Therefore, extracting document-level features is expensive for long scientific articles with a full attention layer. Therefore, we introduce a lightweight attention mechanism inspired by LongFormer \cite{beltagy2020longformer}. The LongFormer language model applies sparse attention between tokens to learn the embedding of the masked tokens. We apply their attention mechanism at the \textbf{sentence level} where each sentence will fully attend locally to the nearby sentences and some sentences will attend globally to all sentences in the document.

\begin{figure*}
\centering
\begin{minipage}{.5\textwidth}
  \centering
  \includegraphics[width=.6\linewidth]{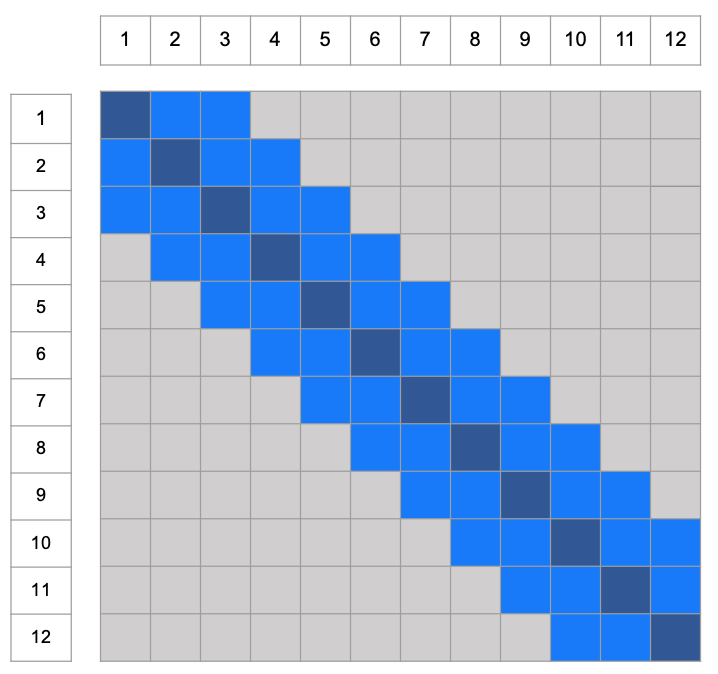}
  \captionof{figure}{Local attention}
  \label{local}
\end{minipage}%
\begin{minipage}{.5\textwidth}
  \centering
  \includegraphics[width=.6\linewidth]{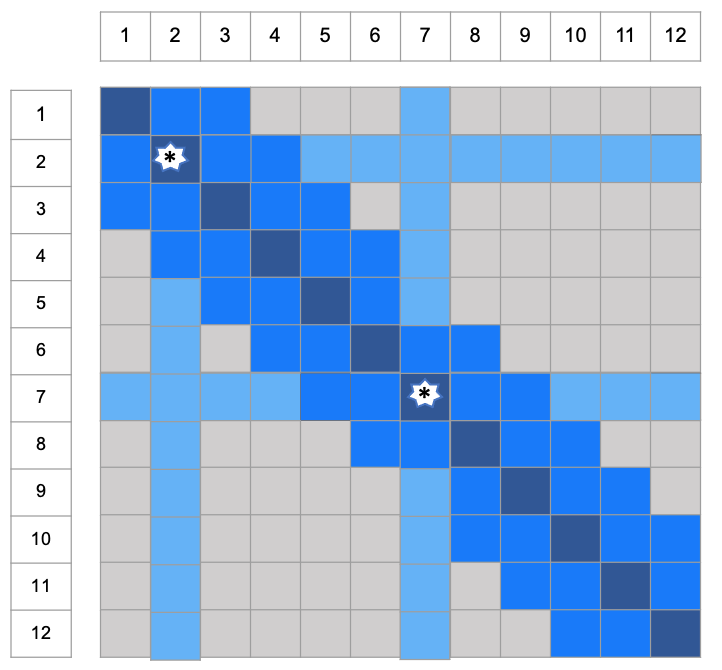}
  \captionof{figure}{Local and global attention} 
  \label{global+local}
\end{minipage}
\label{sparse_attention}
\end{figure*}

This attention mechanism will help model select salient sentences locally from the window and at some random and selected positions sentences will attend to all other sentences to identify the salient sentences that are globally important regardless of the section they belong to. 
The attention window in Figure \ref{local} is 2 which means each sentence will attend to 2 sentences before and after it and in Figure \ref{global+local} sentences 2 and 7 (marked with *) are attending to all other sentences.

Applying window-based local attention requires a few preprocessing steps. We list the main steps in the following section.
\subsubsection{Building the Attention Matrix}

\begin{figure}[ht]
    \centerline{\includegraphics[height=3cm]{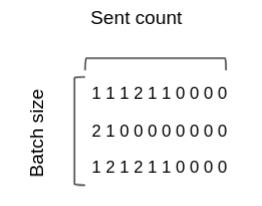}}
    \caption{Attention Mask Matrix}
    \label{attention_mask}
\end{figure}

Since we are processing multiple documents in batch mode and each document has a different length, we fix the number of sentences and make the length of the documents to be a multiple of the attention window. 
Therefore the following steps are required to process the document:
\begin{enumerate}[leftmargin=*]
 \item Padding to document size: the document size is fixed to 500 sentences for scientific documents in our corpus. 
 \item Padding to attention window: the length of the document must be a multiple window size to  be able to apply the sliding window attention mechanism. 
 \item Building attention matrix: the attention matrix has a value of 0 for the padded sentences, a value of 1 for the local attention, and 2 specifies the combination of both local and global attention. Figure \ref{attention_mask} shows an attention matrix for a batch of size 3. This batch contains 3 documents with 6, 2, and  6 sentences respectively. For example, the first document attends locally at positions [1, 2, 3, 4, 5, 6 ] and attends globally at position 4.  
\end{enumerate}
 
\subsubsection{Calculating Attention Value}
Here we list the steps for calculating the local attention. The global attention follows the same approach by adjusting the sentences vectors.

\begin{enumerate}[leftmargin=*]
\item Three linear layers are applied on the sentence vectors to generate the query, key and value vectors

\begin{equation}
    E_q = W_q * E_s + b_q
\end{equation}
\begin{equation}
    E_k = W_k * E_s + b_k
\end{equation}
\begin{equation}
    E_v = W_v * E_s + b_v
\end{equation}

where $E_s$ is the sentence embedding from the embedding layer that embeds the section information. Here $b$ is the bias term and $W$ matrices are learned in the training phase in order to generate $E_q$, $E_k$, and $E_v$ which are respectively the query, key, and value embeddings. 

\item For the second step the query is normalized by the square root of the head dimension
\begin{equation}
    E_q = E_q/ \sqrt{heads}.
\end{equation}
\item The attention scores are calculated by a sliding query and key matrix multiplication on all chunks of the attention-window size 

\begin{equation}
    S_{attn}[i] = E_{q}[i] * E_{v}[i]
\end{equation}

where $E_{q}[i]$ and $E_{v}[i]$ are the query and value embeddings of window $i$, and  $s_{attn}[i]$ is the attention score for that window.

\item The values of the attention scores at the padding positions are set to 0 to ignore values at these locations so that

\begin{equation}
     S_{attn}[padIndex] = 0.
\end{equation}

\item $Softmax$ is applied to attention scores to generate the attention probabilities 
\begin{equation}
     P_{attn} = Softmax(S_{attn}).
\end{equation}

\item  Finally, the attention probabilities are multiplied by the value vectors chunk by chunk in a sliding window 
\begin{equation}
    out[i] = P_{attn}[i] * E_{v}[i].
\end{equation}

\end{enumerate}
\subsection{Transformer Layer}

 Our sparse attention mechanism is applied in each layer of transformation \cite{NIPS2017_7181} and the inputs to the first layer of transformer are the sentence embeddings that include the section information

\begin{equation}
    \widetilde{h}^l = h^{l-1} + normalize( SparseAttention(h^{l-1}))
\end{equation}

\begin{equation}
    h^l =  PositionwiseFeedForward(\widetilde{h}^l)
\end{equation}

where $h^0 = E_{sents}$. We apply a sparse attention mechanism here instead of full attention.

\section{Sentence Extractor}
To generate the final sentence score, we combined the sentence embedding from the language module with a list of features necessary for the score prediction. Section \ref{sent_features} elaborates on the list of features applied to generate the sentence scores. These features depend on the document embedding calculated as in \ref{doc_embed}. 
\subsection{Sentence Features} \label{sent_features}
The features used to predict the final scores are: 
\begin{enumerate}[leftmargin=*]
    \item Length: number of characters in the sentence $i$

    \begin{equation}
    E_{length} = ReLU(Linear(Embedding(length[i]))).
    \end{equation}

    \item Position: position of the sentence ($i$) in the document

    \begin{equation}
    E_{position} = ReLU(Linear(Embedding(i))).
    \end{equation}

    \item Section: section of the sentence $i$ in the document

    \begin{equation}
    E_{section} = ReLU(Linear(Embedding(section[i]))).
    \end{equation}

    Each of the embedding layers is a simple lookup table that stores embeddings of a fixed dictionary and size. The size of the Embedding layers is $d$. The linear layer applies a linear transformation to the input data $x$ ($y = Wx^T + b$).
    \item Correlations:
    the sentence correlations embed the correlation between sentences. The correlation embeddings help the model to identify sentences with a high degree of correlation to other sentences and then exclude them.

    \begin{equation}
    Correlation = tanh( E_{sents} \times W_c \times E_{sents}^T ), 
    \end{equation}

    \begin{equation}
    E_{correlation} = ReLU(Linear(Correlation \times E_{sents}))
    \end{equation}

    \hspace{0.3cm}
    
    where $W_c \in R^{d \times d} $ is the learned correlation matrix and  $Correlation \in R^{n\times n}$ and $E_{Correlation} \in R^{n\times d}$.

    \item Saliency: 
    the saliency embedding will embed the importance of sentence vectors with respect to the document embedding. The saliency weight matrix $W_s \in R^{d \times d}$ is learned in the training phase.

    \begin{equation}
            Saliency = tanh( E_{sents} \times W_s \times E_{D}^T ),
    \end{equation}

    \begin{equation}
    E_{Saliency} = ReLU(Linear(Saliency * E_{sents}))
    \end{equation}

    \hspace{0.3cm}
    
    where $W_s \in R^{d \times d} $ is the learned saliency matrix and $Saliency \in R^{n\times 1}$. The document embedding ($E_D = $) is the weighted average of the sentence embeddings as explained in \ref{doc_embed}
    
\end{enumerate}

\subsection{Document Embedding} \label{doc_embed}

The document encoder is a simply the weighted average of the sentence vectors: 
\begin{equation}
   Weight = Softmax(E_{sents} \times W_{sents})
\end{equation}

where $E_{sents} \in R^{n \times d} $ and $W_{sents} \in R^{d \times 1} $. The wights are initialized randomly and will be learned during the training process. Therefore, $Weight \in R^{n \times 1} $ are the weights of the sentences.

Therefore, the embedding of document $D$ is:

\begin{equation}
    E_{D} = \frac{1}{n} \sum_{i=1}^n Weight[i] * E_{sents[i]}
\end{equation}

where the terms are defined above.

\subsection{Score Predictor} \label{predict}
The score prediction module concatenates all of the features and feeds them to a linear layer to generate the final scores. Our cross-entropy loss evaluates the difference between the prediction and the ground truth scores. We also evaluated the loss factored by the rewards to see if the model makes better predictions using reinforcement learning (in section \ref{rf})

\begin{equation}
\begin{aligned}
p(y_i)= Linear(E_{sent} + E_{length} + E_{position} + \\
E_{section} + E_{correlation} + E_{saliency}) 
\end{aligned}
\end{equation}

where $p(y_i)$ is the probability of adding sentence $i$ to the summary and the linear layer format is $Wx^T+b$ and $x\in R^{1 \times d}$ and $w\in R^{1 \times d}$.

\section{Reinforcement Learning} \label{rf}
Ground truth summaries are abstractive summaries that cover the important content from the input documents. Extracive summarization frameworks need conversion of abstractive summaries to extractive 0/1 labels and they maximize the likelihood of the 0/1 ground truth labels for sentences. 
The Objective function is to minimize this negative log likelihood

\begin{equation}
\label{obj}
    loss = -\sum_{i=1}^{n} log [p(y_i)].
\end{equation}

The objective in Eq. \ref{obj} maximizes the correct prediction of 0/1 labels where $p(y_i)$ is the probability of label $y_i$ for sentence $i$. However, the evaluation of summaries are based on the similarity of selected sentences to the abstractive summaries evaluated by ROUGE scores. Therefore, in the training phase we are minimizing the cross entropy loss while in the test phase we evaluate the ROUGE scores \cite{narayan2018ranking} .  

To mitigate the discrepancy between the train and test objectives, Narayan et. al \cite{narayan2018ranking} suggest using ROUGE scores in the a reinforced setting to factor the pure cross-entropy loss

 \begin{equation}
    \nabla loss \simeq - r(y) \sum_{i=1}^{n} \nabla log [p(y_i)]
\end{equation}

where $r(y)$ is the average of ROUGE-F1 and ROUGE-F2 scores.  

Since there are multiple collection of sentences or candidate summaries that all could have reasonably high ROUGE scores, they suggest training the model with a selection of good candidate summaries. Therefore, if a candidate summary (made by extractive labels) have high overlap with the abstractive summary, the model wants to predict those labels since the $loss$ will be higher.
\section{Experimental Results}

\subsection{Hardware}
Three NVIDIA GPUs (GeForce RTX 2080 Ti) with 11019MiB on memory were used. The batch size is set to 1 because of the size of the input document. Since we could not have a large batch size, we accumulate the gradients for 10 steps and then update the parameters. Learning rate is one of the most important hyper-parameters to tune. We used the NOAM scheduler to adjust the learning rate while training and apply gradient-clipping to prevent exploding gradients. 

\subsection{Experiments}

Table \ref{tune_attn} shows different values for the size of the local attention window and the ratio of the sentences that will attend globally to all other sentences in the document. The results show that increasing the window size for local attention and global attention ratio will improve the ROUGE recall scores. We can set the size of global and local attentions based on the hardware available at hand. Our model converges faster with a larger attention window and more global attentions. 

The effect of reinforcement learning on our model is shown in Table \ref{reinfor_compare}. The reinforcement learning does not improve the results on our dataset mainly due to reducing the bias toward the position and length of the sentences. 

Our results outperform many of the tested extractive and abstractive models as seen in Table \ref{transformer_results}. The sentence scores of BERTSUM in Table \ref{transformer_results} are generated chunk by chunk since this model is not designed for extractive summarization of long documents. The BART and T5 summaries are generated section by section since they were developed for short sequences and have problems with long documents.

 Table \ref{triblockformer} shows the effect of trigram blocking in our dataset. If we block adding a sentence if it has a shared trigram, the results are not improved (first row). We also tried allowing some shared trigram in the sentence. For example, the third row in table \ref{triblockformer} only blocks sentences if there are more than 5 shared trigrams with the current summary. We see that the results get worse with trigram blocking. It shows that the scores predicted by the model are good enough to understand if adding a sentence with shared tokens can improve the ROUGE scores.

\begin{table*}[ht] \setlength{\tabcolsep}{5pt}
\caption[Tuning attention parameters]{Tuning the local attention window size and ratio of global attentions. The evaluation is based on the ROUGE recall scores. The summary limit is 20\% of the size of the input document}
\label{tune_attn}
\centering
\begin{tabular}{c c c c c }
\toprule
  Local Window Size  & Global Ratio(\%) & ROUGE-1  & ROUGE-2  & ROUGE-L \\ 
 \midrule
 \midrule
 6  & - &  56.854 & 19.692 & 41.210 \\
 10 & - & 58.854 & 20.392 & 41.810 \\
 20 & - & 59.06  & 20.77  & 42.00  \\
 30 & - & 58.989 & 20.664 & 42.031 \\
 40 & - & 58.97  & 20.44  & 41.91  \\
 50 & - & 59.408 & 21.099 & 42.232 \\
 \midrule
 40 & 20 & 59.47  & 21.11  & 42.34  \\
 40 & 40 & 59.72  & 21.45  & 42.77  \\
 50 & 20 & \textbf{59.829} & 21.479 & 42.973 \\
 50 & 40 & 59.714 & \textbf{21.498} & \textbf{43.057} \\
 \bottomrule
\end{tabular}
\end{table*}

\begin{table*}[ht] \setlength{\tabcolsep}{5pt}
\caption[Reinforcement Learning Result]{Applying the reinforcement learning suggested in \cite{narayan2018ranking} does not improve ROUGE scores. }
\label{reinfor_compare}
\centering
\begin{tabular}{c c c c c c }
\toprule
  Local Window Size  & Global Ratio(\%)  & Reinforced & ROUGE-1  & ROUGE-2  & ROUGE-L \\ 
 \midrule
 \midrule
 20 & - & No & 59.06  & 20.77  & 42.00  \\
 20 & - & Yes &55.27 &16.40 & 38.54 \\
 \midrule
 30 & - & No & 58.989 & 20.664 & 42.031 \\
 30 & - & Yes & 55.38 & 16.57 & 38.72 \\
 \bottomrule
\end{tabular}
\end{table*}

\begin{table*}[ht] \setlength{\tabcolsep}{5pt}
\caption{Comparison with baselines based on ROUGE recall.}
\label{transformer_results}
\centering
\begin{tabular}{ c c c c }
\toprule
 MODEL & ROUGE-1  & ROUGE-2  & ROUGE-L \\ 
 \midrule
 \midrule
Lead20\% & 37.68 & 6.62 &  15.90  \\ 
TextRank \cite{mihalcea2004textrank} & 38.87 & 9.28 & 19.75 \\
SummaRuNNer \cite{nallapati2017summarunner} & 45.04 & 11.67 & 23.03\\
BART (section-based)  & 46.34 & 11.14 & 29.85 \\
T5 (section-based) & 44.72 & 10.23  & 29.63 \\
BERTSUM & 52.34 & 15.06 & 36.87 \\
SciBERTSUM &  \textbf{59.714} & \textbf{21.498} & \textbf{43.057} \\
 \bottomrule
\end{tabular}
\end{table*}

\begin{table*}[ht] \setlength{\tabcolsep}{4pt}
\caption[Tri-gram blocking negative impact]{ROUGE results for tri-gram blocking.}
\label{triblockformer}
\centering
\begin{tabular}{c c c c c c }
\toprule
  Local Window Size  & Global Ratio(\%)  & Tri-gram count & ROUGE-1  & ROUGE-2  & ROUGE-L \\ 
 \midrule
 \midrule
 
 40 & 40 &  tri-grams \textgreater 0 & 51.27 & 14.57 & 33.34 \\
 40 & 40 &  tri-grams \textgreater 3 & 57.28 & 19.10 & 39.49 \\
 40 & 40 &  tri-grams \textgreater 5 & 58.49 & 20.18 & 40.90 \\
 40 & 40 &  no-blocking   & 59.72 & 21.45 & 42.77 \\
 \bottomrule
\end{tabular}
\end{table*}

\section{Conclusions and Future Work}
We created an extractive summarization framework, SciBERTSUM, based on BERTSUM for long documents with multiple sections (e.g. scientific papers). We generate sentence vectors based on their sections.  The section information is important for the summarization task since sentences in the abstract or method sections are more important compared to the acknowledgement parts. To build a computationally efficient model that scales linearly with the number of sentences in the document, we employed the sparse attention mechanism of LongFormer \cite{beltagy2020longformer} to embed the inter sentence relations. All sentences attend to a limited number of sentences before and after the current sentence and only a small number of random sentences attend globally to all other sentences. Our model is computationally efficient and improves the ROUGE scores on the dataset of paper-slide pairs.

Future work could be applying our model on existing summarization datasets and other long scholarly documents. It would also be interesting to see whether the SciBERT language model, which is pre-trained on scientific text, will give improved performance.

\section{Acknowledgement}
Partial support from the National Science Foundation is gratefully acknowledged.
\bibliographystyle{splncs04}
\bibliography{Sefid}

\end{document}